\documentclass{article}

\usepackage[utf8]{inputenc} 
\usepackage{fullpage}

\usepackage{url}            
\usepackage{booktabs}       
\usepackage{amsfonts}       
\usepackage{nicefrac}       
\usepackage{microtype}      
\usepackage{lipsum}		
\usepackage{graphicx}
\usepackage{doi}

\usepackage{multirow}
\usepackage{multicol}
\usepackage{amsmath}
\usepackage{amssymb}
\usepackage{mathtools}
\usepackage{caption}

\newcommand{\eg}{e.\,g., }
\newcommand{\ie}{i.\,e., }

\title{\textbf{Time-Parameterized Convolutional Neural Networks for Irregularly Sampled Time Series}}

\author{Chrysoula Kosma \\
	\'Ecole Polytechnique, IP Paris\\
	France \\
	\texttt{kosma@lix.polytechnique.fr} \\
	\and
	Giannis Nikolentzos \\
	\'Ecole Polytechnique, IP Paris\\
	France \\
	\texttt{nikolentzos@lix.polytechnique.fr} \\
        \and
	Michalis Vazirgiannis \\
	\'Ecole Polytechnique, IP Paris\\
	France \\
	\texttt{mvazirg@lix.polytechnique.fr} \\
}

\date{}

\begin{document}
\maketitle

\begin{abstract}
Irregularly sampled multivariate time series are ubiquitous in several application domains, leading to sparse, not fully-observed and non-aligned observations across different variables.
Standard sequential neural network architectures, such as recurrent neural networks (RNNs) and convolutional neural networks (CNNs), consider regular spacing between observation times, posing significant challenges to irregular time series modeling.
While most of the proposed architectures incorporate RNN variants to handle irregular time intervals, convolutional neural networks have not been adequately studied in the irregular sampling setting.
In this paper, we parameterize convolutional layers by employing time-explicitly initialized kernels.
Such general functions of time enhance the learning process of continuous-time hidden dynamics and can be efficiently incorporated into convolutional kernel weights.
We, thus, propose the time-parameterized convolutional neural network (TPCNN), which shares similar properties with vanilla convolutions but is carefully designed for irregularly sampled time series.
We evaluate TPCNN on both interpolation and classification tasks involving real-world irregularly sampled multivariate time series datasets.
Our experimental results indicate the competitive performance of the proposed TPCNN model which is also significantly more efficient than other state-of-the-art methods.
At the same time, the proposed architecture allows the interpretability of the input series by leveraging the combination of learnable time functions that improve the network performance in subsequent tasks and expedite the inaugural application of convolutions in this field. 
\end{abstract}

\section{Introduction}\label{Introduction}
Time series arise naturally in many contexts including quantitative finance, astrophysics and medicine, just to name a few.
Recently, there is a growing interest in applying machine learning techniques to time series data.
Besides time series forecasting, which has been extensively studied for decades~\cite{de200625}, other tasks have also emerged recently such as time series classification~\cite{ismail2019deep} and generation~\cite{esteban2017real}.

Time series are constructed from real-world data and usually several of their observations are missing or are subject to noise.
This is mainly due to irregular sampling and is common in different types of data including medical records, network traffic, and astronomical data.
Unfortunately, the most successful machine learning models in sequential modeling, namely recurrent neural networks (RNNs) and convolutional neural networks (CNNs) cannot properly handle such irregularly sampled time series data.
Indeed, those models treat observations successively and assume an equidistant sampling scheme.
Thus, time series data that exhibits variable gaps between consecutive time points pose a significant challenge to such conventional deep learning architectures.
A naive approach to deal with the above problem would be to drop some observations such that the distance between consecutive (remaining) observations is fixed.
However, this would increase data sparsity, thus leading to poorly defined latent variables.
A more prominent approach would be to first apply some imputation method to replace missing values with estimated values, and then to use the standard models which assume an equidistant sampling scheme.
In fact, several recent approaches build on the above idea~\cite{che2018recurrent, futoma2017learning}. 
However, this could potentially result in a loss of information and a violation of the underlying dynamics.

Recently, there has been an increasing interest in effectively capturing the continuous dynamics of real-world sparse and irregular multivariate time series.
Most studies have extended RNNs to continuous-time hidden dynamics defined by ordinary differential equations (ODEs)~\cite{chen2018neural, rubanova2019latent}.
The effectiveness of Convolutional Neural Networks (CNNs)~\cite{lecun1998gradient} as an alternative to recurrent architectures has been established, as long as the input dependencies that are essential fall within the memory horizon of the network.
CNNs are based on parallel computations and thus are more efficient, contrary to the training instability and gradient problems of RNNs that employ back-propagation through time~\cite{werbos1990backpropagation}. 
However, since discrete convolutions learn independent weights for each time step in the kernel range, they do not directly capture the time irregularities.
Efforts for the continuous implementation of convolutional kernels have targeted 3D data~\cite{schutt2017schnet, wang2018deep} and recently, sequences~\cite{romero2021ckconv}. 
The proposed continuous convolution for sequential data~\cite{romero2021ckconv}, CKConv, parameterizes the kernel values using a multi-layer perception (MLP) on the relative positions $\{\Delta_{\tau_i}\}$ of the observations, followed by a periodic activation function~\cite{sitzmann2020implicit}.
In contrast to~\cite{romero2021ckconv} that take advantage of periodic activations, our layer can be constructed employing any predefined set of continuous functions and be followed by any activation, while using significantly fewer learnable parameters, since a single feed-forward layer is used for the parameterization of the convolutional kernel. 

Following the above line of research, in this paper, we develop a new model, so-called \textit{Time-Parameterized Convolutional Neural Network} (TPCNN), which generalizes the standard CNN model to irregularly sampled time series.
To achieve that, we replace the fixed kernels of CNNs with kernels whose values are parameterized both by time and by trainable variables.
Thus, instead of keeping the kernel weights fixed over the whole time series length, we use different functions (e.g., linear, sinusoidal) to produce the kernels that will be convolved with each patch of the time series.
Therefore, kernels can be seen as continuous functions of time, and the proposed TPCNN model can naturally learn continuous latent representations of irregular time series.
Furthermore, the use of the aforementioned functions improves the explainability of the proposed model.
We combine our time-parameterized convolutions with vanilla convolutions by stacking them in a deep encoder module.
The proposed TPCNN model is evaluated in the tasks of time series classification and time series interpolation.
Our experiments demonstrate that the proposed model performs comparably to state-of-the-art methods.
The main contributions of the paper are summarized as follows:
\begin{enumerate}
    \item[(i)] Generalizing conventional, fixed convolutional kernels to time functions, that increase their representational power and still leverage properties of convolutions (e.g., locally aggregated information, fast training).
    \item[(ii)] Enabling the application and proving the efficiency of deep stacked convolutions in the irregular sampling setting.
    \item[(iii)] Achieving high-performance results in interpolation and classification of irregularly sampled benchmark datasets, which are comparable to other state-of-the-art methods.
\end{enumerate}

\section{Related Work}
\label{Related Work}
The long-standing challenge in multivariate irregular time series modeling has led to the development of various neural network architectures that explicitly handle such time-dependent peculiarity. 

One strategy suggests dividing the timeline into equal intervals, filling in missing data, and then using a Recurrent Neural Network (RNN) on the imputed inputs. 
Using a weighted average between the empirical mean and the previous observation to perform imputation has also been proposed~\cite{che2018recurrent}. 
Alternative methods for imputation include the use of Gaussian processes~\cite{futoma2017learning}, or generative adversarial networks~\cite{luo2018multivariate} prior to running the RNN on time-discretized inputs.
The interpolation-prediction network~\cite{shukla2019interpolation} employs several semi-parametric interpolation layers for multivariate time series input with missing values, followed by a prediction network which is applied on the produced regularly spaced and fully observed representations.
Multi-directional RNNs (M-RNN) combine past and future observations for each timestamp~\cite{yoon2018estimating}. 
A differentiable set function method for classifying irregularly sampled is another line of work presented in~\cite{horn2020set}.

An alternative strategy for handling irregularly sampled data involves architectures that directly model such temporal sequences. Various techniques, including adaptations of gated recurrent unit networks (GRUs)~\cite{chung2014empirical} and Long Short-term Memory networks (LSTMs)~\cite{hochreiter1997long}, have been introduced for this purpose.
Among the several proposed modified GRU architectures ~\cite{che2018recurrent}, a prominent example takes as input observed values, indicators denoting missing data points, and the differences in time between observations.
The LSTM architecture has been extended for handling the time irregularity of the data, by introducing a novel time gate in~\cite{neil2016phased} that updates the memory state.
The activation and deactivation of this gate are governed by distinct rhythmic oscillations, controlled by some learnable parameters. 
Another LSTM modification is presented in~\cite{pham2017predicting}, where the proposed forget gate moderates the passing of memory from one time step to another.
Another solution for handling irregularly sampled data is to incorporate the time gaps between observations directly into Recurrent Neural Networks (RNNs). 
One approach is to add the time gap $\Delta_t$ to the RNN input, which has been found to be susceptible to overfitting~\cite{mozer2017discrete}. 
An alternative method is to introduce hidden states that decay over time, which has been proposed in several works as a viable solution~\cite{che2018recurrent, cao2018brits, rajkomar2018scalable}.

Hidden states with an exponential decay can be employed to parameterize neural Hawkes processes and explicitly model observations via latent state changes at each observation event~\cite{mei2017neural}.
Many works focus on the continuous modeling of time series by learning a continuous-time neural representation with a latent state defined at all times.
More specifically, a variational auto-encoder model, which utilizes a neural network decoder in combination with a latent ordinary differential equation (ODE) model, has been presented in~\cite{chen2018neural}. 
Based on this approach, an ODE-RNN encoder that consists of a neural ODE part that models the hidden state dynamics and an RNN part that updates the hidden state has been proposed~\cite{rubanova2019latent}.
A continuous version of the GRU architecture models the input series via continuous ODE dynamics describing the evolution of the probability distribution of the data~\cite{de2019gru}.
Finally, an alternative to Neural ODEs, Neural Controlled Differential Equations represent the continuous-time analogue of an RNN, which benefits from memory-efficient adjoint-based backpropagation across observations~\cite{kidger2020neural}.

Attention mechanisms combined with time encodings, as an alternative to positional ones~\cite{vaswani2017attention}, have been proposed~\cite{song2018attend, zhang2019attain, tan2020data}. 
By extending attention with learnable time embeddings~\cite{xu2019self}, the recently proposed Multi-Time Attention Network~\cite{shukla2021multi} computes the similarity between observations at different time points using a learnable time embedding.
This approach works similarly to kernel-based interpolation, but by leveraging a learnable time attention-based similarity kernel.
Except for the optimization issues of RNNs, the conventional dot-product self-attention mechanism matches queries with keys without considering the surrounding context. At the same time, space complexity grows quadratically with the input length, leading to memory constraints and potential performance limitations.

The use of implicit neural representations for creating continuous data representations by encoding the input in the weights of a neural network has recently gathered interest~\cite{park2019deepsdf,sitzmann2020implicit}.
Our approach can be conceptualized as an implicit representation of the convolutional kernels since they are parameterized as learnable and continuous functions of time.
In this study, the proposed time-parameterized convolutional layer (TPC) introduces time-varying convolutional kernels, allowing for more efficient representational learning of the time dependencies among partially-observed variables. 
We leverage several continuous time functions for extracting learnable time embeddings of the time intervals across different variables.
The proposed architecture is carefully designed for interpolation and classification tasks on irregularly sampled time series.

\section{The TPC Layer}\label{Method}
In this section, we define the mathematical properties of the employed Time-Parameterized layer (TPC) and analytically explain a proposed framework for tasks involving irregularly sampled, partially observed and multivariate time series.

\subsection{Preliminaries}\label{Conv}
Convolution is a well-studied mathematical operation which has applications in many diverse scientific fields~\cite{brigham1988fast}.
The convolution of two functions $f$ and $g$, denoted by $f * g$, expresses how the shape of one is modified by the other.

\paragraph{Continuous convolution.}
If the domains of functions $f$ and $g$ are continuous, convolution is defined as the integral of the product of the two functions after one is reflected and shifted.
Formally, given $f \colon \mathbb{R}^D \rightarrow \mathbb{R}$ and $g \colon \mathbb{R}^D \rightarrow \mathbb{R}$, the continuous convolution operation is defined as: 
\begin{equation*}
    (f \ast g)(\mathbf{x}) = \int_{-\infty}^{\infty}f(\mathbf{y})g(\mathbf{x}-\mathbf{y})d\mathbf{}{y}
\end{equation*}

\paragraph{Discrete convolution.}
In the real world, signals are discrete and finite.
For functions $f$, $g$, defined over the support domain of finite integer set $\mathbb{Z}^D$ and $\{-K, -K+1, ..., K-1, K\}^D$, respectively, the discrete equivalent of convolution is defined as:
\begin{equation}
    (f \ast g)[n] = \sum_{k=-K}^{K}f[n-k]g[k]
    \label{eq:conv}
\end{equation}
Thus, the integral is replaced by a finite summation.
Standard CNN models consist of layers that perform discrete convolutions that are defined over the discrete domain.

\subsection{Time-Parameterized 1D Convolutions}
\label{TPCNN}
We first introduce the key notations behind the employed time-parameterized convolutions for irregular and multivariate time series and analyze their fundamental properties.

\paragraph{Irregular time series and standard CNNs.}
Let $\{\mathbf{X}^{(1)}, \ldots, \mathbf{X}^{(N)}\}$ be a collection of multivariate time series where $\mathbf{X}^{(i)} \in \mathbb{R}^{m \times L}$ for all $i \in \{1,\ldots,N\}$.
Thus, each time series consists of $m$ channels and has a length (i.e., number of observations) equal to $L$ which corresponds to the observation times $\{t_1, t_2, \ldots, t_L\}$.
Let also $d(\cdot, \cdot)$ denote a function that measures the distance (in time) between observations of a single channel of the collection of time series.
The convolution operation of standard CNNs assumes that consecutive observations are equally spaced across all samples, and thus, the weights of the different kernels of standard CNNs are fixed across all chunks of the time series.
In other words, the summation in the right part of Equation~\eqref{eq:conv} is performed over the elements of the same set for all $n$.
Formally, we have that $d \Big(\mathbf{X}_{i,j}^{(\imath)}, \mathbf{X}_{i,j+1}^{(\jmath)} \Big) = \tau$ holds for all $i \in \{1,\ldots,m \}$, $j \in \{ 1,\ldots, L-1\}$ and $i, j \in \{1,\ldots,N\}$ where $N$ is the number of samples.
However, the above does not necessarily hold in the case of irregularly sampled time series data.
Indeed, irregular sampling for multivariate series leads to variations in the number of observations across channels.
Thus, due to the assumptions it makes, the standard convolution operation of CNNs is not suitable for irregular time series data.

\paragraph{Time-parameterized convolutional kernels.}
To deal with the irregularity of time series, we propose to use time-parameterized kernels.
Thus, instead of a fixed kernel that slides over the patches of the time series, we use a parameterized kernel whose components are functions of time.
The kernel is also parameterized by the weights of a neural network.
We constraint the size of the kernel to be equal to $2z+1$ where $z \in \mathbb{N}_0$ where $\mathbb{N}_0$ denotes the set of natural numbers together with zero.
Then, the elements of the kernel are constructed by some function $g(\theta, \Delta t)$ where $\theta$ denotes some trainable parameters and $\Delta t$ denotes the distance (in time) of the observation associated with some element of the kernel and the $z+1$-th observation.
Formally, the convolution is defined as follows:
\begin{equation}
    (f \ast g)(t) = \sum_{i=1}^{2z+1} f(t_i) g(\theta, t-t_i) = \sum_{i=1}^{2z+1} f(t_i) g(\theta, \Delta t_i)
\end{equation}
where $t_1,\ldots,t_{2z+1}$ are the timestamps associated with the observations of the patch the kernel is applied to.

The function $g(\theta, \Delta t)$ is quite general and can have different forms.
In this paper, we focus on interpretability and thus function $g(\theta, \Delta t) \colon \mathbb{R}^5 \rightarrow \mathbb{R}$ is defined as follows:
\begin{equation*}
    g \Big( \begin{bmatrix} \theta_1 & \theta_2  & \theta_3 & \theta_4 & \Delta t \end{bmatrix}^\top \Big) = \theta_1\bigg( \sigma \Big( h\big(\theta_3 \cdot \Delta t + \theta_4\big) \Big) + \theta_2 \bigg)
\end{equation*}
where $h \colon \mathbb{R} \rightarrow \mathbb{R}$ is a continuous function in $\mathbb{R}$ and $\sigma \colon \mathbb{R} \rightarrow \mathbb{R}$ denotes some activation function (i.e., sigmoid, ReLU, etc.).
Thus, to construct each element of the kernel, function $g$ takes as input four trainable parameters (i.e., $\theta_1,\theta_2,\theta_3$ and $\theta_4$) and the time difference between the current observation and the center observation of the patch. 
Function $h$ is chosen such that inductive bias is injected into the model.
This can allow the model to capture patterns that commonly occur in time series data and also make its internal operations more interpretable.
For example, a function $h(x) = c$ where $c$ is some constant would not be a good candidate for extracting useful features from the time series.
On the other hand, we employ more informative functions which can capture useful properties of time series such as trend and seasonality.
In particular, we employ the following ten functions:
\begin{multicols}{2}
    \begin{enumerate}
        \item $h_1(x) = x$
        \item $h_2(x) = \sin(x)$
        \item $h_3(x) = \cos(x)$
        \item $h_4(x) = \tan(x)$
        \item $h_5(x) = \exp(x)$
        \item $h_6(x) = x^2$
        \item $h_7(x) = x^3$
        \item $h_8(x) = \sinh(x)$
        \item $h_9(x) = \cosh(x)$
        \item $h_{10}(x) = \tanh(x)$
    \end{enumerate}
\end{multicols}
Most of the time, trend is a monotonic function, and therefore, functions $h_1, h_6$ and $h_7$ are chosen to detect trend in time series.
Seasonality is a typical characteristic of time series in which the data experiences regular and predictable changes that recur over a defined cycle.
Functions $h_2, h_3, h_9$ and $h_{10}$ are responsible for extracting features that take seasonality into account.

The approach presented above generates kernels for univariate time series.
In the case of multivariate time series, different parameters are learned for the different components of the time series.
Therefore, the four parameters ($\theta_1, \theta_2, \theta_3$ and $\theta_4$) are replaced by vectors of dimension $m$, \ie $\boldsymbol{\theta}_1, \boldsymbol{\theta}_2, \boldsymbol{\theta}_3, \boldsymbol{\theta}_4 \in \mathbb{R}^m$.
Thus, function $g(\boldsymbol{\theta}, \Delta t) \colon \mathbb{R}^{4m+1} \rightarrow \mathbb{R}^m$ is computed by applying function $h(\cdot)$ pointwise to $m$ different elements.
Note that $\Delta t$ is still a scalar since observation times are identical across all components of the series.





\subsection{The Time-Parameterized Convolutional (TPC) Layer}
Given a sample $\mathbf{X}^{(i)}$, its corresponding observation times $\{t_1, t_2, \ldots, t_L\}$, and a time-parameterized function $g$, the kernel centered at the $j$-th observation (\ie $\mathbf{X}_{:,j}^{(i)}$) is constructed as follows:
\begin{center}
\small
\begin{tabular}{lccccc} \hline
    \textbf{Patch} & $\mathbf{X}_{:,j-K}^{(i)}$ & \ldots & $\mathbf{X}_{:,j}^{(i)}$ & \ldots & $\mathbf{X}_{:,j+K}^{(i)}$ \\ \hline
    \textbf{Observation time} & $t_{j-K}$ & \ldots & $t_j$ & \ldots & $t_{j+K}$ \\ \hline 
    \textbf{Difference in time} &  $\Delta t_{j-K}$ & \ldots & $0$ & \ldots & $\Delta t_{j+K}$ \\ \hline
    \textbf{Kernel} & $g(\boldsymbol{\theta}, \Delta t_{j-K})$ & \ldots & $g(\boldsymbol{\theta}, 0)$ & \ldots & $g(\boldsymbol{\theta}, \Delta t_{j+K})$ \\ \hline
\end{tabular}
\end{center}
Note that $\mathbf{X}_{:,j}^{(i)}$ denotes the $j$-th column of matrix $\mathbf{X}^{(i)}$.
Once we construct the kernel, the output of the convolution is computed as follows:
\begin{equation*}
    \begin{split}
        c = \sum_{l=1}^m g(\boldsymbol{\theta}, \Delta t_{j-K})_l \, \mathbf{X}_{l,j-K}^{(i)} + \ldots &+ \sum_{l=1}^m g(\boldsymbol{\theta}, 0)_l \, \mathbf{X}_{l,j}^{(i)} + \ldots \\
        &+ \sum_{l=1}^m g(\boldsymbol{\theta}, \Delta t_{j+K})_l \, \mathbf{X}_{l,j+K}^{(i)}
    \end{split}
\end{equation*}
where $c \in \mathbb{R}$.
In some cases, features of the multivariate time series might be missing.
In such cases, the above operation would compute the sum of a smaller number of terms (since missing features are ignored).
Thus, we also experimented with the mean function:
\begin{equation}
    \label{eq:mean}
    \begin{split}
    c = \frac{1}{\nu} \Bigg( \sum_{l=1}^m g(\boldsymbol{\theta}, \Delta t_{j-K})_l \, \mathbf{X}_{l,j-K}^{(i)} + \ldots &+ \sum_{l=1}^m g(\boldsymbol{\theta}, 0)_l \, \mathbf{X}_{l,j}^{(i)} + \ldots \\
    &+ \sum_{l=1}^m g(\boldsymbol{\theta}, \Delta t_{j+K})_l \, \mathbf{X}_{l,j+K}^{(i)} \Bigg)
    \end{split}
\end{equation}
where $\nu$ denotes the actual number of features (out of the $(2K+1)m$ features, those that are not missing).

Thus, the convolution between a sequence of observations and the kernel outputs a real number.
We use zero padding and apply the kernel to all observations and, therefore we obtain a vector $\mathbf{c} \in \mathbb{R}^L$.
Furthermore, similar to standard CNNs, not a single kernel, but instead a collection of kernels is generated and applied to the input.
These kernels might correspond to different functions of the ones defined above (\ie $h_1, \ldots, h_{10}$).
Suppose that we use $p$ different kernels in total (potentially of different functions).
Then, the output of the TPC layer of the multivariate and irregularly sampled time series $\mathbf{X}^{(i)}$ is computed as:
\begin{equation*}
    TPC(\mathbf{X}^{(i)}, \mathbf{t}^{(i)}) = \big\Vert^p_{i=1} \mathbf{c}_i \in \mathbb{R}^{L \times p}
\end{equation*}
where $\Vert$ is the concatenation operator between vectors and $\mathbf{t}^{(i)}$ is a vector that stores the observation times of the time series.


\subsection{Properties of TPC Layer}\label{TPCNN-properties}

\paragraph{Constant number of parameters}
An interesting property of the TPC layer is that the number of parameters of each kernel is constant and equal to $4m$ regardless of the size of the kernel.
This is because the kernel is dynamically generated based on the observation times and only $4m$ trainable parameters are involved.
This is in contrast to standard convolutional layers where the number of parameters is equal to the size of the kernel plus the bias.
Thus, the number of parameters of the TPC layer will be less than the number of parameters of a standard convolutional layer when the size of the kernels is greater than $4$.
This is likely to lead to less complex models and might significantly reduce overfitting.

\paragraph{Time Complexity.}
The time complexity of the proposed TPC layer is approximately $\mathcal{O}(L \ell m p)$ for kernel size $\ell$, similar to the vanilla 1D convolution. 
Since TPC relies on convolutions, that take advantage of parallel computations, it can be trained faster than recurrent neural network architectures.
The complexity comparison becomes even more advantageous when compared with continuous-time models, such as neural ODEs that are significantly slower than RNNs~\cite{kidger2020neural}.


\paragraph{Learning Properties.}
The proposed TCP layer introduces time-varying convolutional kernels as opposed to fixed kernels that are commonly employed in traditional convolutional neural networks (CNNs). 
In other words, the employed kernels do not remain fixed throughout the whole length of the input series. 
This particular trait of TPC does not explicitly force weight sharing between different subsequences of the time series during convolution. 
Weight sharing is, however, implicitly modeled via the learnable representations of time, that are used to initialize the kernel weights.
This is based on the assumption that observations that are mapped to similar time embeddings will probably share similar values of weights in the convolutional operation.
The proposed approach still maintains the ability to locally aggregate information by retaining the notion of fixed kernel size in the convolution operation.
This allows for the output of the convolution to be locally aggregated, while still incorporating the benefits of time-varying convolutional kernels.

\paragraph{Invariance Properties.}
If some patterns in the time series are identical, both in terms of the observations and also in terms of the difference in time between the observations, then the TPC layer will produce the same output for those two patterns.
For example, let $\mathbf{x}_i = (x_{i-K}, \ldots, x_{i}, \ldots, x_{i+K})$ and $\mathbf{x}_j = (x_{j-K}, \ldots, x_{j}, \ldots, x_{j+K})$ denote two sequences of values and $\mathbf{t}_i = (t_{i-K}, \ldots, t_{i}, \ldots, t_{i+K})$ and $\mathbf{t}_j = (t_{j-K}, \ldots, t_{j}, \ldots, t_{j+K})$ denote their respective observation times.
If $\mathbf{x}_i = \mathbf{x}_j$ holds and $\Delta \mathbf{t}_i = \Delta \mathbf{t}_j$ also holds, where $\Delta \mathbf{t}_i = (t_{i-K} - t_i, \ldots, 0, \ldots, t_{i+K} - t_i)$ and $\Delta \mathbf{t}_j = (t_{j-K} - t_j, \ldots, 0, \ldots, t_{j+K} - t_j)$, then the kernels produced for these two sequences of values are identical and therefore, the layer produces the same output.

Furthermore, the different functions defined in the previous subsection make the kernels invariant to different transformations.
For instance, in the above example, suppose that $\Delta \mathbf{t}_i \neq \Delta \mathbf{t}_j$, and that the $k$-th element of the second sequence is equal to $(k+1)2\pi$ times the corresponding element of the first sequence for $k \in \{0,1,\ldots,2K+1\}$.
Then, the TPC layer equipped with the $h_2$ function (\ie $\sin(\cdot)$ function) and with $\theta_3=1$ and $\theta_4=0$ would produce the same output for both patterns.
Such a function can capture periodic temporal correlations.


\begin{figure*}
\centering
  \includegraphics[width=0.8\textwidth]{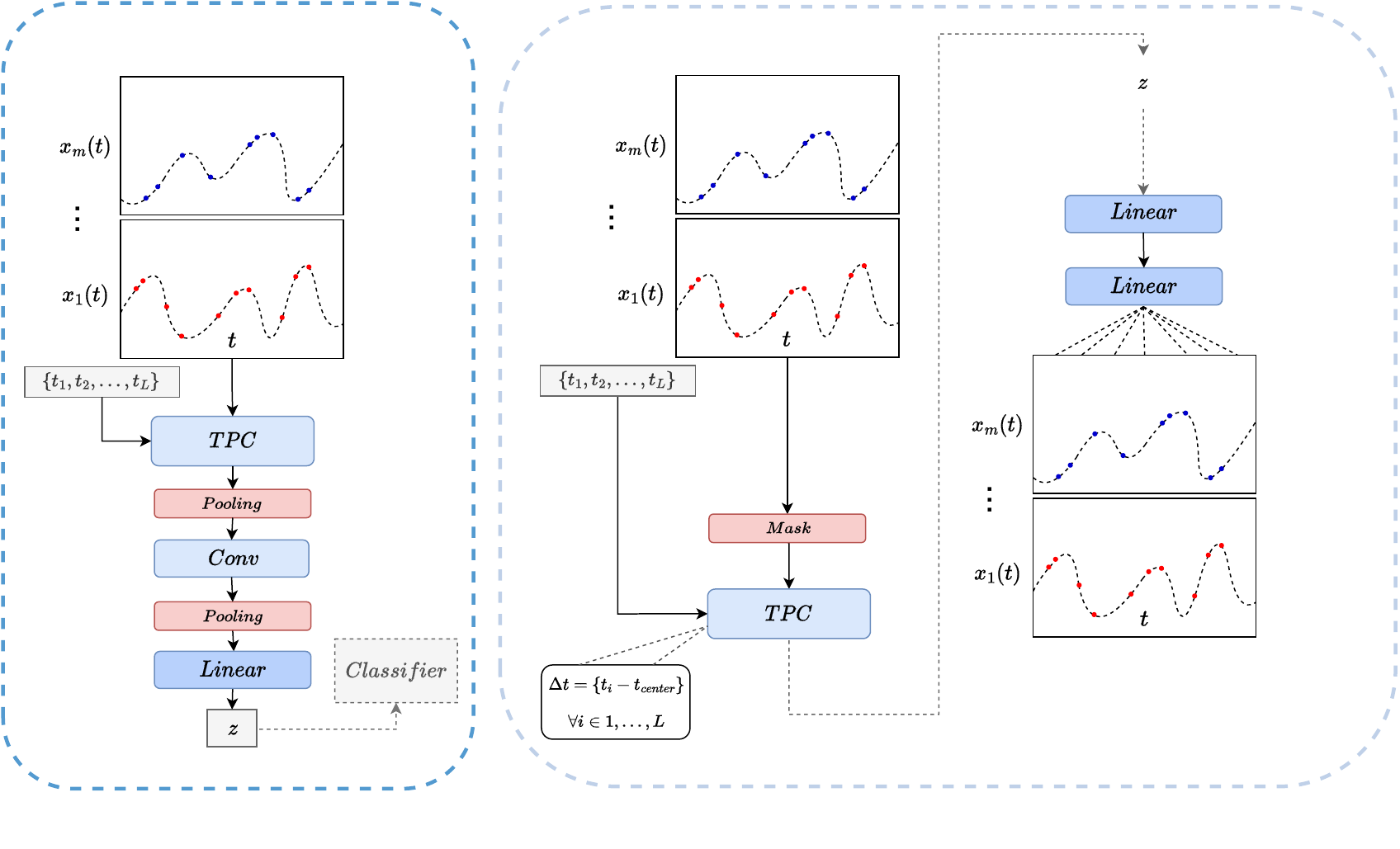}
  \caption{(Left) An encoder that consists of the proposed TPC layer, convolutions and pooling layer and produces a fixed-size latent representation $\boldsymbol{z}$. (Right) An encoder-decoder framework that reconstructs the series from the input using TPC and linear layers. }
\label{fig:pconv-enc-dec}
\end{figure*}

\subsection{TPCNN Framework for Irregularly Sampled Time Series}
\label{TPCNN-enc-dec}
We will next discuss how the TPC layer can be integrated into neural network architectures for dealing with various tasks that involve irregular time series, such as interpolation and classification.
Following previous work, we propose an encoder-decoder framework, so-called Time-Parameterized Convolutional Neural Network (TPCNN) framework.
In what follows, we give more details about the two main components of the proposed framework, namely its encoder and its decoder.

\paragraph{TPCNN Encoder.}
This module is responsible for mapping the input time series into a latent vector which captures their overall shape and their specificities.
The first layer of the encoder is an instance of the TPC layer introduced above.
The TPC layer receives as input the irregular and multivariate series $\mathbf{X}^{(i)} \in \mathbb{R}^{m \times L}$ and the corresponding vector of observation times $\mathbf{t}^{(i)} = \{t_1, t_2, ..., t_L\}$.
The output of TPC layer is then successively fed to vanilla convolution layers which can capture longer-time dependencies of the continuous latent representation of the time series. 
A pooling layer follows each convolution layer, including TPC.
By down-sampling the output, such layers are expected to extract features that are good indicators of class membership or of the shape of the time series.
Finally, a fully-connected layer is applied to the output of the last convolution layer to extract a low-dimensional representation $\mathbf{z}^{(i)} \in \mathbb{R}^d$ of the series.

\paragraph{TPCNN Decoder.}
This part of the architecture is responsible for reconstructing the multivariate input series from the latent vector that is produced by the encoder.
The latent vector $\mathbf{z}$ that was produced by the encoder is first given as input to a fully-connected layer whose objective is to perform rescaling.
The emerging vector is then passed onto another fully-connected layer which produces a matrix $\hat{\mathbf{X}}^{{(i)}}$ that matches the dimension of the input time series.
These reconstructed time series are then compared against the input series to evaluate the autoencoder's performance.

\paragraph{Interpolation and Classification Setting.}
Note that some components of the TPCNN framework depend on the considered task, \ie interpolation or classification.
For instance, in the interpolation setting, each time a kernel of the TPC layer is applied to some subset of the input series, the observation that lies at the center of that subset is masked such that the model does not have direct access to it.
On the other hand, such a masking is not performed in the case of the classification setting.

The reconstruction loss of a standard autoencoder is typically measured using the mean squared error (MSE) between the original input and the reconstructed output.
For an input time series $\mathbf{X}^{{(i)}}$, the MSE loss is computed as:
\begin{equation*}
    \mathcal{L}_{interpolation} = \frac{1}{|O|}\sum_{j \in \mathcal{O}} {\|\mathbf{X}_{:,j}^{(i)} - \hat{\mathbf{X}}_{:,j}^{(i)} \|}^2_2
\end{equation*}
where $\mathcal{O}$ is a set that contains the indices of the observed values and $\hat{\mathbf{X}}^{(i)}$ denotes the reconstructed series produced by the decoder as a function of the latent vector $\mathbf{z}$.

The encoder-decoder framework of Figure~\ref{fig:pconv-enc-dec} (Right) is combined with the MSE loss for the interpolation task. 
Additionally, as already discussed, masking is performed on the center element of each slice of the input series, and the rest of the observed values of the slice are used for interpolation.

In the case of classification, the latent representation $\mathbf{z}$ that is generated by the encoder and which preserves  the information about the multivariate time series' dependencies, can be directly fed to a classifier module to make predictions.
In the experiments that follow, we employ a 2-layer multi-layer perceptron (MLP) with $ReLU$ activation function.
Thus, in the case of a classification problem with $|\mathcal{C}|$ classes, the output is computed as follows:
\begin{equation*}
    \mathbf{p} = softmax(MLP(\mathbf{z}))
\end{equation*}
Then, given a training set consisting of time series $\mathbf{X}^{(1)}, \ldots, \mathbf{X}^{(N)}$, we use the negative
log-likelihood of the correct labels as training loss:
\begin{equation*}
    \mathcal{L}_{classification} = -\sum_{i=1}^N \sum_{j=1}^{|\mathcal{C}|} \mathbf{y}_j^{(i)} \log \mathbf{p}_j^{(i)}
\end{equation*}
where $\mathbf{y}_j^{(i)}$ is equal to $1$ if $\mathbf{X}^{(i)}$ belongs to the $j$-th class, and $0$ otherwise.

The application of the TPCNN model to the above two scenarios is illustrated in Figure \ref{fig:pconv-enc-dec} (classification on the left and interpolation on the right).

\section{Experiments}
\label{Experiments}
In this section, we describe the experimental setup and methodology used to evaluate the performance of our proposed time-parameterized convolutional layer on various tasks involving irregular time series, including interpolation and classification.

\subsection{Datasets}
\label{Datasets}
We evaluate the performance of the proposed architecture and the baselines on the following real-world datasets: 

\paragraph{PhysioNet:}
The PhysioNet Challenge $2012$ dataset~\cite{silva2012predicting} comprises $8000$ multivariate time series that correspond to records from the first $48$ hours of a patient's admission to the intensive care unit (ICU). Measurements include $37$ variables which can be missing at different steps and occur in irregular intervals.
Half of the instances are labeled with $13.8\%$ of instances being in the positive class (in-hospital mortality). 
For the interpolation experiments, we used all 8000 instances and for the classification experiments, we used the 4000 labeled instances. 
We use the same experimental protocols and preprocessing steps as in~\cite{rubanova2019latent}.

\paragraph{MIMIC-III:}
The MIMIC-III dataset~\cite{johnson2016mimic} consists of multivariate health records, that can have missing values, collected at Beth Israel Deaconess Medical Center between $2001$ and $2012$. 
Based again on the preprocessing strategy of~\cite{rubanova2019latent}, we extract $53211$ samples including $12$ features.
Given the first $48$ hours of data, the task is to predict in-hospital mortality, with $8.1\%$ of the data samples in the positive class.

\paragraph{Human Activity:}
The human activity dataset contains time series data from five individuals performing various activities (such as walking, sitting, lying, standing, etc.), based on the 3D positions of tags attached to their belts, chest and ankles ($12$ features in total).
Following the preprocessing procedures outlined by~\cite{rubanova2019latent}, a dataset of $6554$ sequences and $50$ time steps is extracted. 
The task for this dataset is to classify each time step in the series into one of the eleven activities. 

\subsection{Experimental Setting}
\label{experimental-setting}
We next explain the experimental setting we follow for interpolation and classification, similar to the work of~\cite{shukla2021multi}.
In the case of interpolation, we study all instances (labeled and unlabeled) from the PhysioNet dataset. The dataset is partitioned into an $80\%$ training set and a $20\%$ test set, with a fraction ($20\%$) of the training data serving as the validation set. The interpolation task is to predict based on a subset of available data points values for the unobserved points. This is executed using different percentages of observed steps, which vary between $50\%$ and $90\%$ of the total available steps.  
For this experiment, we perform five different runs and report performance on the unobserved data using the mean squared error (MSE) metric.

We also use the labeled data from the PhysioNet, MIMIC-III and Human Activity datasets to conduct classification experiments. 
For the physiological data of PhysioNet and MIMIC-III, the classification task considers the entire time series, whereas, in the context of the human activity dataset, classification is performed for each time step in the series.
We follow the same train, validation and test splitting procedure as described in the interpolation setting.
For this experiment, we perform five different runs to provide the classification performance on the different datasets.
For PhysioNet and MIMIC-III datasets, we report performance using the area under the ROC curve (AUC) score, due to class imbalance. 
For the Human Activity dataset, we asses the model performance using the accuracy metric.
The validation set is used to select the best set of hyperparameters for our models via grid search. 

\begin{table*}[t]
\caption{Performance for interpolation with different percentages of observed time points on \textit{PhysioNet}. We mention in bold the best-performing method(s) and underline the second best-performing method(s) based on statistical significance tests.}
\label{interpolation-results}
\begin{center}
\begin{small}
\begin{sc}
\begin{tabular}{lccccc}
\toprule
\textbf{Model} & \multicolumn{5}{c}{\textbf{Mean Squared Error} ($\times 10^{-3}$)} \\
\midrule
RNN-VAE & 13.418 $\pm$ 0.008 & 12.594 $\pm$ 0.004 & 11.887 $\pm$ 0.007 & 11.133 $\pm$ 0.007 & 11.470 $\pm$ 0.006 \\
L-ODE-RNN & 8.132 $\pm$ 0.020 & 8.140 $\pm$ 0.018 & 8.171 $\pm$ 0.030 & 8.143 $\pm$ 0.025 & 8.402 $\pm$ 0.022\\
L-ODE-ODE & 6.721 $\pm$ 0.109 & 6.816 $\pm$ 0.045 & 6.798 $\pm$ 0.143 & 6.850 $\pm$ 0.066 & 7.142 $\pm$ 0.066\\
mTAND-Full & \textbf{4.139 $\pm$ 0.029} & \textbf{4.018 $\pm$ 0.048} & \textbf{4.157 $\pm$ 0.053} & \textbf{4.410 $\pm$ 0.149} & \textbf{4.798 $\pm$ 0.036}\\
TPCNN (ours) &\underline{5.993 $\pm$ 0.058} &\underline{5.797 $\pm$ 0.063} & \underline{5.654 $\pm$ 0.108} & \underline{5.624 $\pm$ 0.084} & \underline{5.532 $\pm$ 0.140} \\
\midrule
\textbf{Observed}(\%) & 50\% & 60\% & 70\% & 80\% & 90\% \\ 
\bottomrule
\end{tabular}
\end{sc}
\end{small}
\end{center}
\end{table*}

\subsection{Baseline Models}
\label{Models}
In this study, we conduct a thorough evaluation of several deep learning architectures as baseline models for performance comparison. 
These models are specifically designed to handle irregular time series and include variations of the Recurrent Neural Network (RNN), Attention modules and encoder-decoder architectures. 

The specific models evaluated in this study include:
\begin{enumerate}
  \item[(i)] Basic RNN variants including: \textit{RNN-Impute}, \textit{RNN-}$\Delta_t$, \textit{RNN-decay}, \textit{GRU-D}. 
  The \textit{RNN-Impute} model employs a method to impute missing data points based on the weighted average between the last observation of the time series and the total mean of the variable in the training set~\cite{che2018recurrent}. 
  In \textit{RNN-}$\Delta_t$ the input to RNN is extended with a missing indicator for the variable and the time interval $\Delta_t$ since the last observed point.
  The \textit{RNN-decay} is an RNN with hidden states that decay exponentially over time~\cite{mozer2017discrete, che2018recurrent}, whereas \textit{GRU-D} employs exponential decay on both hidden states and input~\cite{che2018recurrent}.
  \item[(ii)] Other RNN variants, such as \textit{Phased-LSTM}, \textit{IP-Nets}, \textit{SeFT}, \textit{RNN-VAE}. 
  The \textit{Phased-LSTM} model incorporates time irregularity through the use of a time gate that controls access to the hidden and cell states of the LSTM~\cite{neil2016phased}. 
  \textit{IP-Nets} are Interpolation-Prediction Networks (IPN), which perform interpolation prior to prediction with an RNN on the transformed equally-spaced intervals, using semi-parametric interpolation layers~\cite{shukla2019interpolation}.
  The \textit{SeFT} model employs learnable set functions for time series and combines the representations with an attention-based mechanism~\cite{horn2020set}.
  \textit{RNN-VAE} is a standard variational RNN encoder-decoder.
  \item[(iii)] ODE variants, such as \textit{ODE-RNN}, \textit{L-ODE-RNN}, \textit{L-ODE-ODE}.
  In \textit{ODE-RNN} neural ODEs model the dynamics of the hidden state, and an RNN updates the hidden state in the presence of new observations~\cite{rubanova2019latent}.
  Similarly, \textit{L-ODE-RNN} and \textit{L-ODE-ODE} are latent ODEs with the former combining an RNN encoder and a neural ODE decoder~\cite{chen2018neural}, and the latter an ODE-RNN encoder and a neural ODE decoder~\cite{rubanova2019latent}.
  \item[(iv)] Attention-based frameworks, including \textit{mTAND}. 
  The multi-time attention network, \textit{mTAND}, interpolates missing data using a learnable attention similarity kernel between observations, which are accessed based on trainable temporal embeddings~\cite{shukla2021multi}.
\end{enumerate}

\subsection{Results}

\paragraph{Interpolation of missing data.} 
In Table~\ref{interpolation-results} we present the results of the experimental setting designed for interpolation, as described in Section~\ref{experimental-setting}. 
For different percentages of observed values (\ie ranging from $50\%$ to $90\%$), we record the interpolation performance on the reconstructed irregularly sampled multivariate time series of the PhysioNet dataset using the MSE metric.
We compare the proposed TPCNN model to different baseline methods designed for interpolation, including RNN-VAE, L-ODE-RNN, L-ODE-ODE and mTAND-Full (\ie mTAND encoder-decoder framework for interpolation). 
We mention in bold the best-performing method and underline the results for the second-performing method.
We also perform tests for measuring the statistical significance of the studied methods, which leads to highlighting two distinct models that refer to the highest performances.
We can observe that the best-performing method is mTAND-Full, which is closely followed by the proposed TPCNN model. 
The rest of the baselines show significantly worse performance compared to the proposed TPCNN, including the highly accurate in the irregular setting ODE-based method L-ODE-ODE.
The performance of the proposed model ranges from $\sim 6.0 \times 10^{-3}l$ to $\sim 5.5 \times 10^{-3}$ in terms of MSE, showing a slightly improved performance as the percentage of missing observations decreases.
On the other hand, mTAND-Full shows a slightly degrading performance for a smaller percentage of missing data, with RNN-VAE being the only baseline method that follows the same behavior.

\begin{table*}[t]
\caption{Performance for \textbf{per-sequence} classification on \textit{PhysioNet} and \textit{MIMIC-III} and \textbf{per-time-point} classification on \textit{Human Activity} datasets. We mention in bold the best-performing method(s) and underline the second best-performing method(s) based on statistical significance tests.}
\label{classification-results}
\begin{center}
\begin{small}
\begin{sc}
\begin{tabular}{lccc}
\toprule
\multirow{2}{*}{\textbf{Model}} & \multicolumn{2}{c}{\textbf{\underline{AUC}}} & \underline{\textbf{Accuracy}} \\
                        & \textbf{PhysioNet} & \textbf{MIMIC-III} & \textbf{Human Activity} \\
\midrule
RNN-Impute & 0.764 $\pm$ 0.016 & 0.8249 $\pm$ 0.0010 & 0.859 $\pm$ 0.004 \\
RNN-$\Delta_t$ & 0.787 $\pm$ 0.014 & 0.8364 $\pm$ 0.0011 & 0.857 $\pm$ 0.002 \\
RNN-Decay & 0.807 $\pm$ 0.003 & 0.8392 $\pm$ 0.0012 & 0.860 $\pm$ 0.005 \\ 
RNN GRU-D & 0.818 $\pm$ 0.008 & 0.8270 $\pm$ 0.0010 & 0.862 $\pm$ 0.005 \\
Phased-LSTM & \underline{0.836 $\pm$ 0.003} & 0.8429 $\pm$ 0.0035 & 0.855 $\pm$ 0.005 \\
IP-Nets & 0.819 $\pm$ 0.006 &  0.8390 $\pm$ 0.0011 & 0.869 $\pm$ 0.007 \\
SeFT & 0.795 $\pm$ 0.015 &  \underline{0.8485 $\pm$ 0.0022} & 0.815 $\pm$ 0.002 \\
RNN-VAE & 0.515 $\pm$ 0.040 & 0.5175 $\pm$ 0.0312 & 0.343 $\pm$ 0.040 \\
ODE-RNN & \underline{0.833 $\pm$ 0.009} & \textbf{0.8561 $\pm$ 0.0051} & 0.885 $\pm$ 0.008 \\
L-ODE-RNN & 0.781 $\pm$ 0.018 & 0.7734 $\pm$ 0.0030 & 0.838 $\pm$ 0.004 \\
L-ODE-ODE & \underline{0.829 $\pm$ 0.004} & \textbf{0.8559 $\pm$ 0.0041} & 0.870 $\pm$ 0.028 \\
mTAND-Full & \textbf{0.858 $\pm$ 0.004} & \textbf{0.8544 $\pm$ 0.0024} & \textbf{0.910 $\pm$ 0.002} \\
\midrule
TPCNN (ours) & \underline{0.833 $\pm$ 0.001} &0.8380 $\pm$ 0.0011  & \underline{0.897 $\pm$ 0.004} \\
\bottomrule
\end{tabular}
\end{sc}
\end{small}
\end{center}
\end{table*}

\begin{table*}[t!]
\caption{Memory and computational costs, in terms of size (number of parameters) and time per epoch (in minutes).}
\label{model-size}
\begin{center}
\begin{small}
\begin{sc}
\def\arraystretch{1.3}
\begin{tabular}{lccc}
\toprule
\textbf{Model} & \textbf{PhysioNet} &\textbf{MIMIC-III} & \textbf{Human Activity}  \\
\midrule
 & \multicolumn{3}{c}{Size (parameters)} \\
mTAND-Full &1.3M  &1.4M  &1.6M  \\
TPCNN (ours) &350K  &100K  &300K   \\ 
\midrule
& \multicolumn{3}{c}{Time per epoch $(min)$} \\
mTAND-Full &0.06  &0.5  &0.006  \\
TPCNN (ours) &0.15  &0.2  &0.008   \\
\bottomrule
\end{tabular}
\end{sc}
\end{small}
\end{center}
\end{table*}

\paragraph{Classification.}
We also report in Table~\ref{classification-results} the results of the different baselines, as described in Section~\ref{Models}, and the proposed TPCNN model on classification for the labeled instances of PhysioNet, MIMIC-III and Human Activity datasets.
For the first two imbalanced datasets, we use AUC as an evaluation metric and perform per-sequence binary classification, whereas, for the Human Activity dataset, we report accuracy for the task of per-time-point classification.
For all datasets, we boldly mention the best-performing methods and underline the results for the second best-performing methods.
Due to several non-statistically significant differences in performances, we have several methods being among the first or second best-performing.
For PhysioNet and Human Activity datasets, our proposed TPCNN framework is the second-best method in terms of metrics, surpassed by the attention-based model mTAND-Full.
More specifically, in PhysioNet the proposed model performs as well as the ODE variants (\ie ODE-RNN, L-ODE-ODE) that are however significantly slow in terms of computational time, as mentioned in~\cite{shukla2021multi}.
In Human Activity classification, TPCNN shows quite improved performance being $\sim 1\%$ worse than mTAND-Full.
However, in the MIMIC-III classification, the proposed TPCNN model lies among the third-best-performing methods, being surpassed by several baselines. 
In this dataset, ODE-RNN, L-ODE-ODE and mTAND-Full methods achieve the highest AUC scores, followed by the SeFT model, which however performs significantly worse in classification experiments for the other two datasets.
The significant performance advantage of mTAND-Full in this task can be attributed to its design which jointly performs interpolation and classification while directly attending only to observed time points.
On the other hand, the proposed model handles missing data inside the convolutional kernel of the TPC layer by applying the mean aggregator of Equation~\ref{eq:mean}.
The aggregation neighborhood however is constrained by the kernel size and remains fixed throughout the series length.
Extending the proposed architecture to incorporate size-varying kernels could further improve the learning capabilities of the TPC layer.

\begin{figure*}[t]
\centering
  \includegraphics[width=0.8\textwidth]{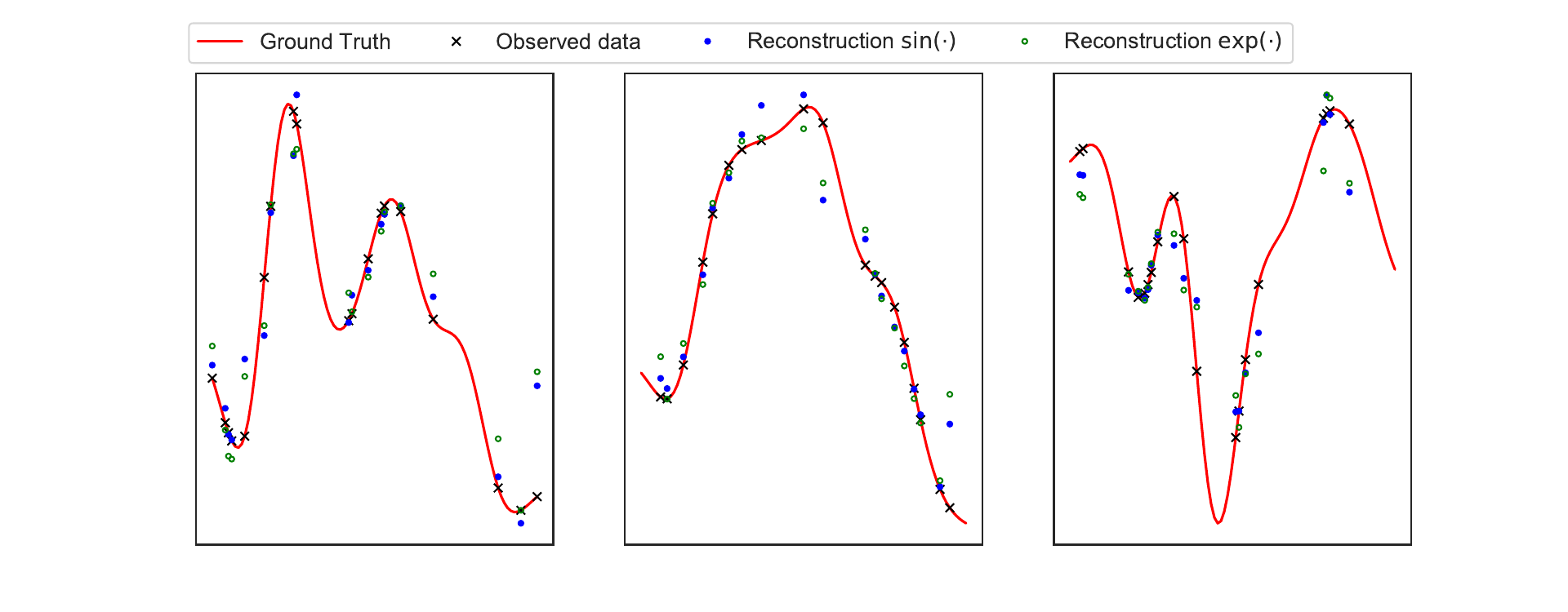}
  \caption{Reconstruction results using the proposed TPCNN model on the synthetic dataset. Three different samples of the test set are visualized.}
\label{fig:synthetic_reconstruction}
\end{figure*}

\paragraph{Computational cost.}
In Table~\ref{model-size} we provide a comparison in terms of memory and computational costs between the proposed TPCNN and its main competitor mTAND-Full. 
We report the size, \ie the number of parameters, and the time per epoch in minutes for the two methods and the three real-world datasets.
Comparisons of mTAND and previous state-of-the-art models, among which the efficient ODE-based methods, as shown in~\cite{shukla2021multi} have demonstrated that the former is significantly faster (\ie approximately 100 times) than ODE-based methods that make use of an ODE solver.
As we can observe in Table~\ref{model-size}, TPCNN is as fast as mTAND-Full in terms of time cost comparison.
When it comes to the size of the model, the proposed TPCNN uses significantly fewer parameters compared to mTAND-Full, while maintaining competitive performance.
More specifically, TPCNN uses approximately some hundred thousand parameters, \ie $100-350$ thousand parameters, while mTAND-Full size scales to millions of parameters, \ie approximately $1.5$ million. 
This comparison highlights the high efficacy of convolutions in the irregular sampling setting, which allow the training of neural networks that are significantly smaller and fast compared to the baselines.
Therefore, the proposed TPCNN can easily scale to larger datasets and remains efficient even when trained with fewer parameters.

\paragraph{Experiments on synthetic data.}
Following the line of work of~\cite{shukla2021multi}, we reproduce their synthetic sinusoidal dataset that consists of $1000$ samples, each describing a time series of $100$ time points where $t \in [0,1]$.
Given $10$ reference points, an RBF kernel with bandwidth $100$ is used to obtain local interpolations at the $100$ time steps.
For each sample, $20$ time points are randomly selected so as to represent an irregularly spaced series.
A split of $80\%$ and $20\%$ extracts the respective train and test sets.
We employ the encoder-decoder interpolation framework of Figure~\ref{fig:pconv-enc-dec} (Right).
Contrary to the interpolation setting for PhysioNet, we give as input the $20$ irregular time steps, without the missing points, and reconstruct each observation based on the rest using TPCNN with the functions $h_2(x) = \sin(x)$ (blue points) and $h_5(x) = \exp(x)$ (green points).
We visualize the obtained reconstructions for $3$ samples of the test set in Figure~\ref{fig:synthetic_reconstruction}.
Each plot consists of the true values (ground truth) for a test sample, while the dark markers represent the $20$ observed input data points (observed data), the blue markers and the green markers the $20$ predicted values (reconstruction) using $\sin(\cdot)$ and $\exp(\cdot)$ functions respectively.
By employing the function $h_2(x) = \sin(x)$, we are able to achieve a lower MSE loss compared to the ones achieved with the rest of the time functions defined in Section~\ref{TPCNN}.
We should mention here that in case domain knowledge is available, it can be incorporated into the proposed TPCNN method via the employed time function, which is likely to lead to performance improvements.

\begin{figure*}[t!]
\centering
  \begin{minipage}{.33\linewidth}
  \centering
  {\includegraphics[width=\linewidth]{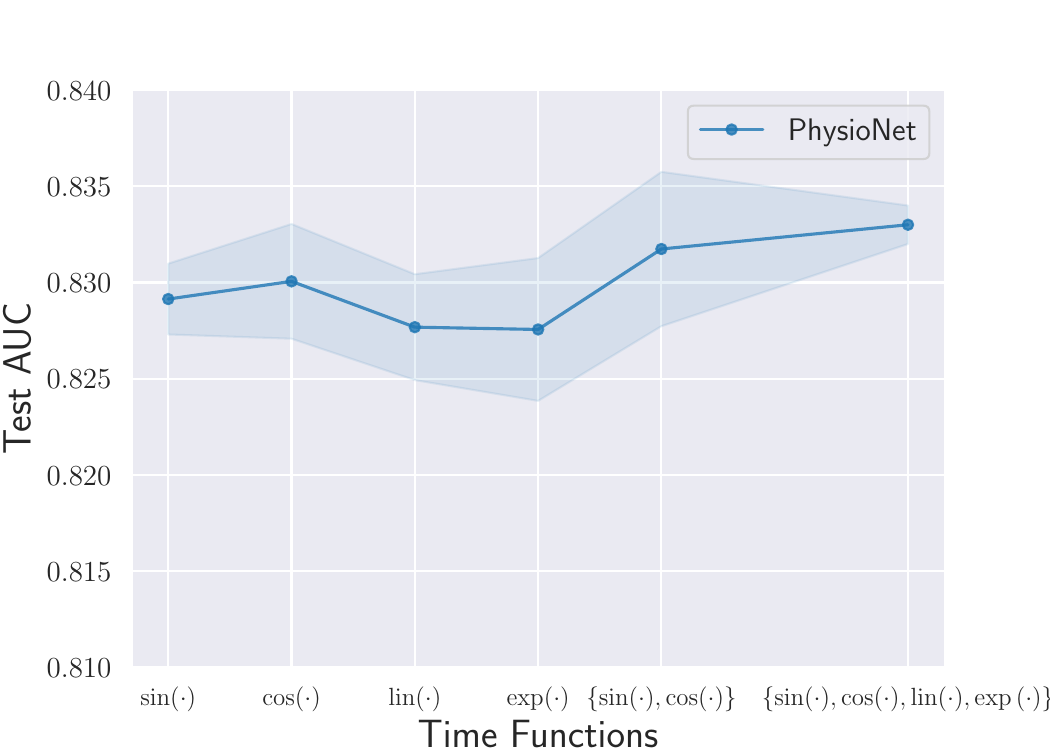}}
  \end{minipage}
  \begin{minipage}{.33\linewidth}
  \centering
  {\includegraphics[width=\linewidth]{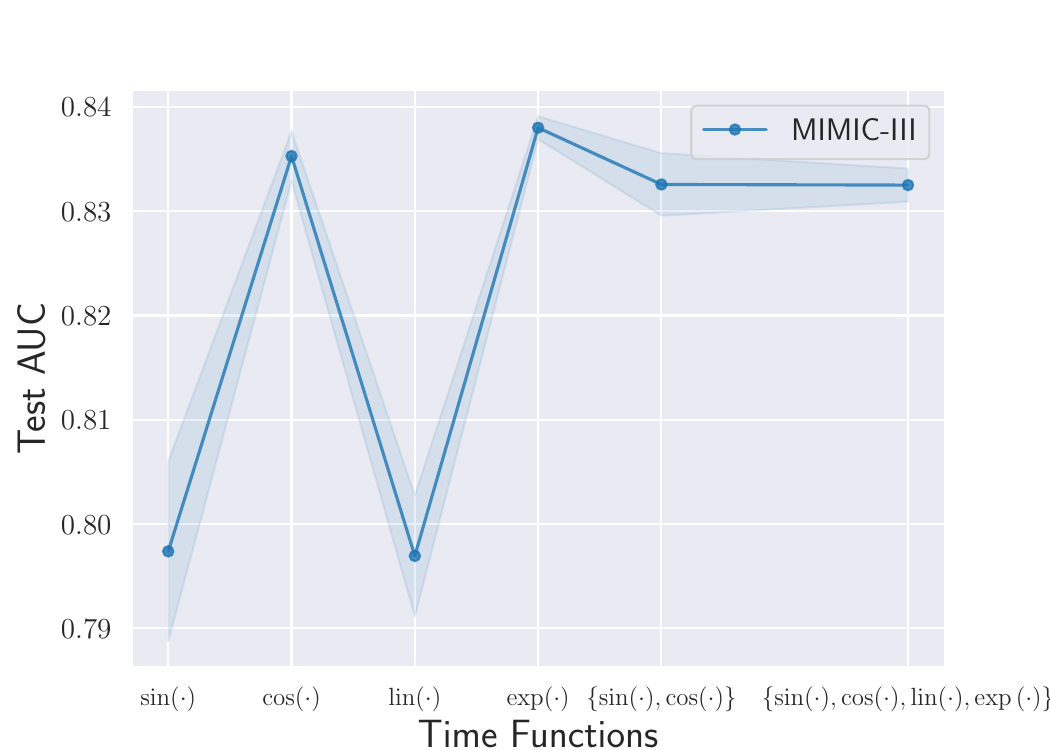}}
  \end{minipage}
  \begin{minipage}{.33\linewidth}
  \centering
  {\includegraphics[width=\linewidth]{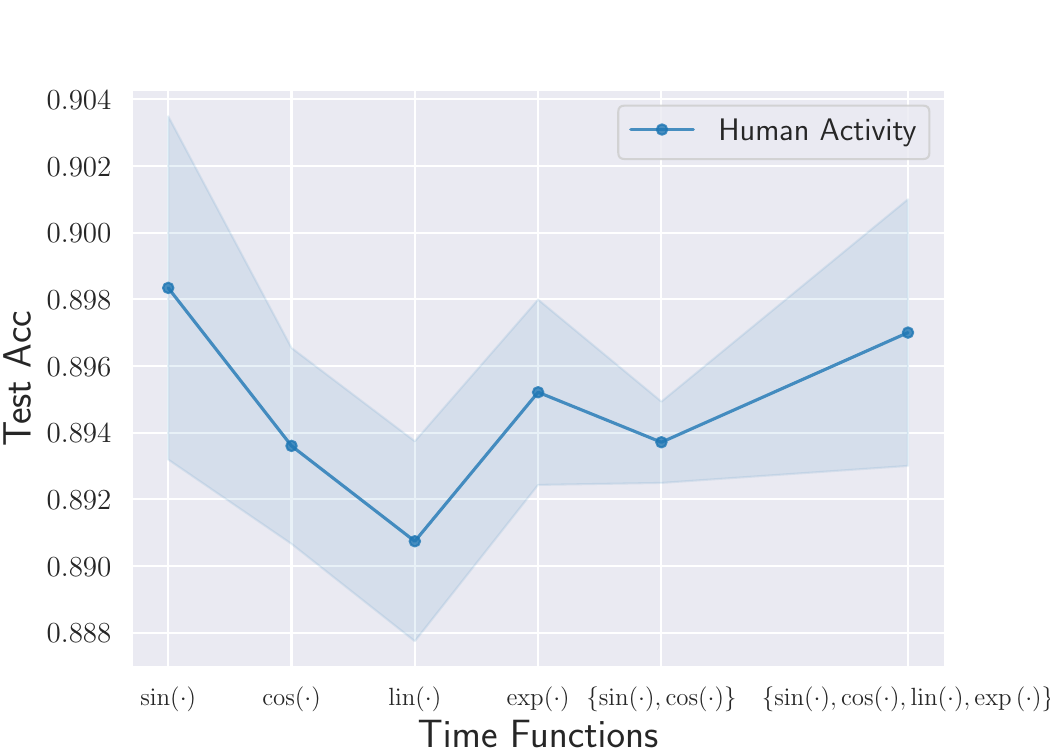}}
  \end{minipage}
  \caption{Ablation study on different time functions for the parameterization of convolutional kernels for each dataset. Each plot captures the performance (AUC or Accuracy) for each function or combination of functions on the test set.}
\label{fig:ablation}
\end{figure*}

\paragraph{Ablation study.}
We also present in Figure~\ref{fig:ablation} an ablation study on different time functions employed for parameterizing the weights of the convolutional kernels.
The performance metric (AUC or accuracy) on the test set is reported on the classification task of the real-world datasets given a different time function or combination of time functions.
For all three datasets, we examine a subset of the functions described in Section~\ref{TPCNN}.
More specifically, we employ $h_1(x),h_2(x),h_3(x),h_5(x)$ (\ie $\text{lin}(\cdot),\sin(\cdot),\cos(\cdot),\exp(\cdot)$) and their combination (\eg $\{\sin(\cdot),\cos(\cdot)\}, \{\sin(\cdot),\cos(\cdot),\text{lin}(\cdot),\exp(\cdot)\}$).
We observe that different functions may contribute more or less to the classification performance for the given dataset.
In PhysioNet, while the linear function $\text{lin}(\cdot)$ and exponential function $\exp(\cdot)$ lead to the lowest AUC on the test set, when combined with $\sin(\cdot)$ and $\cos(\cdot)$ they achieve a performance improvement by $\sim 1 \%$.
Additionally, in MIMIC-III classification $\cos(\cdot)$ and $\exp(\cdot)$ functions show the highest AUC test, while $\sin(\cdot)$ and $\text{lin}(\cdot)$ (\ie linear function) lead to a reduced performance by $\sim 4 \%$.
At the same, the combination of functions improves performance but does not surpass $\cos(\cdot)$ and $\exp(\cdot)$ when employed alone.
Finally on the Human Activity dataset, $\cos(\cdot)$ function and the combination $\{\sin(\cdot),\cos(\cdot),\text{lin}(\cdot),\exp(\cdot)\}$, followed by the $\exp(\cdot)$ function achieve the highest test accuracy.
The linear $\text{lin}(\cdot)$ function again, in this case, leads to the lowest accuracy score compared to the rest of the time functions.
During training, we can observe that the linear time function followed by a standard non-linear activation (\eg ReLU) when used for the parameterization of the convolutional kernel weights suffers from slow convergence and consequently worse performance.
On the other hand, periodic time functions and the exponential function seem to more efficiently describe the time dynamics and lead to smoother training when used for parameterizing convolutions.
This experiment highlights the explainability aspects of the proposed TPCNN model since it allows us to determine which time functions better describe the considered time series.
Furthermore, under certain conditions, the time series could be considered as a composition of such kind of functions.

\section{Conclusion}
\label{Conclusion}
In this work, we carefully designed and experimentally evaluated a novel time-parameterized convolutional neural network, which incorporates learnable time functions into the weights of convolutional kernels.
The proposed method generalizes well in different tasks involving irregularly sampled multivariate time series while being computationally efficient and interpretable.

\bibliographystyle{plain}
\bibliography{arxiv}

\end{document}